\documentclass[conference]{IEEEtran}
\IEEEoverridecommandlockouts

\usepackage{amsmath,amssymb,amsfonts}
\usepackage{pgfplots}
\usepackage{pgfplotstable}
\pgfplotsset{compat=1.16}
\usepackage{graphicx}
\usepackage{xcolor}
\usepackage{multirow}
\usepackage{booktabs}
\usepackage{rotating}
\usepackage{algorithm}
\usepackage{algpseudocode}
\usepackage{pifont}
\usepackage{makecell}
\usepackage{hyperref}
\usepackage{colortbl}
\usepackage[font=small,justification=justified,singlelinecheck=false]{caption}

\hypersetup{
    colorlinks=true,
    linkcolor=blue,
    filecolor=magenta,
    urlcolor=cyan,
    pdftitle={Dynamic Temperature Scheduler for Knowledge Distillation},
    pdfpagemode=FullScreen,
}

\usepackage[shortlabels]{enumitem}
\newcommand{\cmark}{\ding{51}} % check mark
\newcommand{\xmark}{\ding{55}} % cross mark
\begin{document}
	
	\title{Dynamic Temperature Scheduler for Knowledge Distillation.}
	
	\author{\IEEEauthorblockN{Sibgat Ul Islam\textsuperscript{1}, Jawad Ibn Ahad\textsuperscript{1}, Fuad Rahman\textsuperscript{2}, Mohammad Ruhul Amin\textsuperscript{3}, Nabeel Mohammed\textsuperscript{1}, Shafin Rahman\textsuperscript{1}}
		
		\IEEEauthorblockA{\textsuperscript{1}Apurba-NSU R\&D Lab, Department of Electrical and Computer Engineering, \\ North South University, Dhaka, Bangladesh}
		\IEEEauthorblockA{\textsuperscript{2}Apurba Technologies, Sunnyvale, CA 94085, USA}
		\IEEEauthorblockA{\textsuperscript{3}Fordham University, New York, USA}
		\textsuperscript{1}\{sibgat.islam, jawad.ibn, nabeel.mohammed, shafin.rahman\}@northsouth.edu \\
		\textsuperscript{2}fuad@apurbatech.com, \textsuperscript{3}mamin17@fordham.edu\\
	}
	
	\maketitle
	
	\begin{abstract}
		Knowledge Distillation (KD) trains a smaller student model using a large, pre-trained teacher model, with temperature as a key hyperparameter controlling the softness of output probabilities. Traditional methods use a fixed temperature throughout training, which is suboptimal. Moreover, architectural differences between teacher and student often result in mismatched logit magnitudes. We demonstrate that students benefit from softer probabilities early in training but require sharper probabilities in later stages. We introduce Dynamic Temperature Scheduler (DTS), which adjusts temperature dynamically based on the cross-entropy loss gap between teacher and student. To our knowledge, this is the first temperature scheduling method that adapts based on the divergence between teacher and student distributions. Our method integrates seamlessly with existing KD frameworks. We validate DTS across multiple KD strategies on vision (CIFAR-100, Tiny-ImageNet) and NLP tasks (GLUE, Dolly, SelfIns, UnNI, S-NI), consistently outperforming static-temperature baselines. Code is available at \url{https://github.com/Sibgat-Ul/DTS}.
	\end{abstract}
	
	\begin{IEEEkeywords}
		Knowledge Distillation, Temperature Scheduling
		\vspace{-0.5em}
	\end{IEEEkeywords}
	
	\section{Introduction}
	
	\noindent Knowledge Distillation (KD), introduced by Hinton et al.~\cite{kd}, is a powerful technique for transferring knowledge from a large teacher model to a smaller student model. This technique involves softening the output probabilities of both the teacher and student using a temperature hyperparameter. A lower temperature sharpens the probability distribution, encouraging the student to focus on the largest logits, while a higher temperature softens the distribution and exposes the student to "dark knowledge'' by emphasizing the negative logits. The distillation process typically utilizes Kullback-Leibler (KL) divergence with temperature scaling, enabling the student to learn this dark knowledge. Traditionally, the temperature is determined through extensive grid search and may vary depending on the dataset, specific instances within the dataset, and differences in model architecture and complexity. Most research aimed at improving the KD process can be categorized into three main areas: \textbf{(a) Logits Distillation:} Transfers knowledge by matching the temperature-softened output probabilities \cite{kd, DKD}. \textbf{(b) Feature Distillation:} Aligns intermediate feature representations  between teacher and student. \textbf{(c) Relational Distillation:} Preserves relationships between samples or layers rather than individual output features~\cite{li2022knowledge, huang2022knowledge}. Despite these advancements, the optimization of the temperature parameter, a key component of KD, has received relatively little attention. Works such as RLKD~\cite{RLKD}, CTKD~\cite{CTKD}, AKD~\cite{AKD}, and metaKD~\cite{metaKD} that optimize temperature require additional modules or training steps.
	
	\begin{table}[t]
		\centering
		\caption{Comparison of temperature adjustment methods. \textcolor{green}{Green} indicates favorable characteristics (e.g., no learnable modules), while \textcolor{red}{Red} indicates less favorable characteristics (e.g., additional modules, multi-stage training).}
		\begin{tabular}{lcccc}
			\toprule
			\textbf{Criteria} & \textbf{AKD} & \textbf{CTKD} & \textbf{DTS (Ours)} \\
			\midrule
			Learnable module     & \textcolor{green}{\xmark} & \textcolor{red}{\cmark} & \textcolor{green}{\xmark} \\
			Multi stage training & \textcolor{red}{\cmark} & \textcolor{green}{\xmark} & \textcolor{green}{\xmark} \\
			Dynamic              & \textcolor{red}{\xmark} & \textcolor{green}{\cmark} & \textcolor{green}{\cmark} \\
			Extra Hyperparameter      & \textcolor{red}{\cmark} & \textcolor{red}{\cmark} & \textcolor{red}{\cmark} \\
			Learning method       & \textcolor{green}{\xmark} & \textcolor{red}{adversarial} & \textcolor{green}{\xmark} \\
			% Compatibility        &  partial & partial & full \\
			\bottomrule
		\end{tabular}
		\vspace{-0.5em}
	\end{table}
	
	Through this work, we are introducing our novel Dynamic Temperature Scheduler (DTS). The scheduler has three main hyperparameters: \textit{max temperature}, \textit{min temperature}, \textit{initial temperature}. The training starts from a higher initial temperature that decays as the training progresses to the minimum temperature. This scheduling process mimics the curriculum strategy of CTKD. Additionally, during the scheduling process, DTS dynamically adjusts the temperature with respect to the difference in cross-entropy loss between the teacher and the student's output with the true labels.
	
	Our contributions in this work can be summarized as follows: (1) We show that a static temperature is not always ideal for the KD process. (2) We show that different temperatures might be required in different phases of training, as the student learns the teacher's distribution. (3) A novel Dynamic Temperature Scheduler can be seamlessly integrated into existing KD frameworks that helps the student adapt to the teacher.
	
	\section{Related Works}
	\label{sec: rel_work}
	\noindent\textbf{Knowledge Distillation.} The KD proposed by Hinton et al. \cite{kd} aims to transfer knowledge from a pre-trained teacher model to a student model through logits softened by a fixed temperature in logit-based methods. A higher temperature will soften the teacher's output more and a lower temperature produces sharper logits. However, the reason for setting this crucial hyperparameter fixed was not discussed.
	
	\noindent\textbf{Temperature Adjustment in KD.} To address the issue of having different temperatures, \textbf{AKD} \cite{AKD} proposes a two-stage training strategy: (1) annealing temperature from \( \mathcal{T}_{\text{max}} \) to \( \mathcal{T} = 1 \) using mean squared error (MSE), (2) fine-tuning at \( \mathcal{T} = 1 \) using cross-entropy (CE) loss with the ground truth labels. Our DTS does not require multiple stages and adjusts the temperature dynamically. The \textbf{CTKD} \cite{CTKD} has two learnable temperature modules Global-T and Instance-T to predict the temperature. CTKD also includes a parameter \(\lambda\) to create a curriculum learning strategy. In contrast, our DTS requires no additional learnable modules or MLPs and comes with curriculum learning. \textbf{RLKD} \cite{RLKD} provides flexible and adaptive scheduling formulating temperature adjustment, where an agent observes the temperature to maximize a reward on improving student performance. Unlike ours, it introduces computational overhead due to the complexity of the Reinforcement Learning pipeline.
	
	\section{Methodology}
	\noindent Consider a teacher model \( f_T \) and a student model \( f_S \), both trained on a dataset \( \mathcal{D} = \{(\mathbf{x}_n, y_n)\}_{n=1}^N \), where each input \( \mathbf{x}_n \in \mathbb{R}^{H \times W} \) (e.g. an image) and \( y_n \in [1, C] \) is the corresponding class label. For any given input \( \mathbf{x}_n \), the teacher and the student produce logits \( \mathbf{v}_n = f_T(\mathbf{x}_n) \) and \( \mathbf{z}_n = f_S(\mathbf{x}_n) \), respectively. These logits are converted to softened probability distributions via temperature-scaled softmax:
	\begin{align*}
		\mathbf{P}_T^{(k)}(\mathcal{T}) &= \frac{\exp(v_n^{(k)} / \mathcal{T})}{\sum_{m=1}^C \exp(v_n^{(m)} / \mathcal{T})}, \\
		\mathbf{P}_S^{(k)}(\mathcal{T}) &= \frac{\exp(z_n^{(k)} / \mathcal{T})}{\sum_{m=1}^C \exp(z_n^{(m)} / \mathcal{T})}
		\label{eq:softmax}
	\end{align*}
	\indent where \( \mathcal{T} \) is the temperature parameter, and \( v_n^{(k)}, z_n^{(k)} \) denote the logits corresponding to class \( k \).
	\subsection{Problem Formulation}
	\begin{figure}[!t]
		\centering
		\includegraphics[scale=0.7]{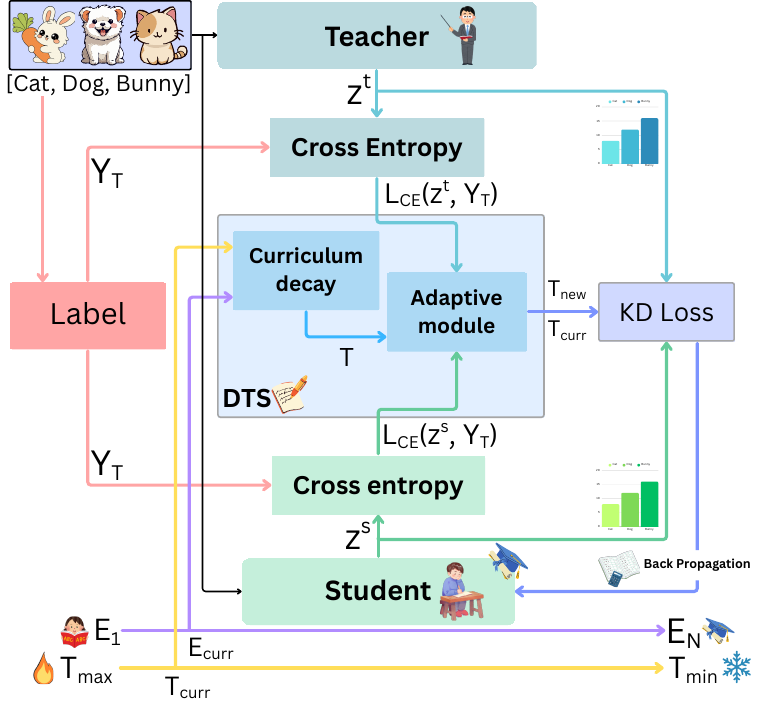}
		
		\caption{A high level overview of our Dynamic Temperature Scheduler (DTS). The logits from the models are used to calculate cross-entropy loss with the true labels then passed to the DTS along with the temperature at current epoch. Using these a new temperature is calculated which is then used for the distillation.}
		
		\label{fig:overview}
		\vspace{-0.5em}
	\end{figure}
	
	\noindent The temperature \( T \) in the softmax function controls the smoothness of the resulting probability distribution:
	\[
	P_i = \frac{\exp(z_i / \mathcal{T})}{\sum_{j=1}^C \exp(z_j / \mathcal{T})},
	\]
	\noindent\textbf{High \( \mathcal{T} \)}: Produces a softer, more uniform distribution by compressing logits. \textbf{Low \( \mathcal{T} \)}: Yields a sharper, more confident distribution by amplifying logit differences. In the context of KD \cite{kd}, the student model is trained to mimic the output distribution of the teacher. The distillation loss \cite{kd} is typically defined using the Kullback-Leibler divergence:
	\[
	\mathcal{L}_{\text{KD}} = \mathcal{T}^2 \cdot \sum_{i=1}^C P_{\text{T}, i}(\mathcal{T}) \cdot \log\left(\frac{P_{\text{T}, i}(\mathcal{T})}{P_{\text{S}, i}(\mathcal{T})}\right),
	\]
	\indent where the \( \mathcal{T}^2 \) factor ensures that the gradient magnitudes are appropriately scaled during backpropagation. The gradient of this loss with respect to the student’s logits is:
	\[
	\frac{\partial \mathcal{L}_{\text{KD}}}{\partial z_i^{\text{S}}} \propto \frac{1}{\mathcal{T}} \left(P_{\text{S}, i}(\mathcal{T}) - P_{\text{T}, i}(\mathcal{T})\right).
	\]
	\indent This highlights a crucial insight: temperature directly modulates the learning signal. Larger \( \mathcal{T} \) results in gentler gradients; ideal for early training, while smaller \( \mathcal{T} \) emphasizes fine-grained distinctions useful in later stages.
	\noindent\textbf{Limitations of a static temperature.} A static temperature assumes that the optimal softening level remains constant throughout the training. This assumption proves problematic for two key reasons: \textbf{(a) Early training:} The predictions of the student model are uncertain and produce smaller logits. A higher temperature helps extract relative class relationships from the teacher's distribution. \textbf{(b) Late training:} As the student becomes more confident, its logits become more discriminative. A lower temperature is necessary to encourage sharper distinctions.
	
	\subsection{Solution Strategy}
	\label{sec: sol_strat}
	\noindent The gradients derived from the KD loss scale inversely with temperature. Therefore, (a) Too low a temperature early in training can result in overly aggressive updates, potentially destabilizing learning, and (b) Too high temperature later on may lead to vanishing gradients.
	
	\noindent\textbf{(a) Too low temperature early on can yield overly aggressive updates.} At the beginning of training, the student's logits are typically close to zero: \( z_j^{\text{S}} \approx 0 \). The softmax at temperature \( \mathcal{T} \) becomes:
	\[
	P_{\text{S}, i}(\mathcal{T}) = \frac{e^{z_i / \mathcal{T}}}{\sum_{j=1}^C e^{z_j / \mathcal{T}}} \approx \frac{1}{C}, \quad \text{if } z_i \approx 0.
	\]
	\indent The KD loss gradient with respect to the student logits is given by
	\[
	\frac{\partial \mathcal{L}_{\text{KD}}}{\partial z_i^{\text{S}}} = \frac{1}{\mathcal{T}} \left( P_{\text{S}, i}(\mathcal{T}) - P_{\text{T}, i}(\mathcal{T}) \right).
	\]
	\indent When \( \mathcal{T} \to 0 \), the term \( \frac{1}{\mathcal{T}} \) becomes large, magnifying even small discrepancies between teacher and student distributions. This leads to large gradients, resulting in: (a) Overshooting updates during backpropagation, and (b) Destabilized early training and poor convergence. Thus, using a low temperature early in training amplifies sensitivity to initial distribution mismatches and harms stability.
	\begin{table}[!t]
		\centering
		\caption{Top-1 Accuracy (\%) of different KD methods on CIFAR-100. Highest accuracies are in \textbf{bold}.}
		\label{tab:diff_scheduler}
		\begin{tabular}{cccc}
			\toprule
			\textbf{Teacher} & \textbf{Student} & \textbf{Method} & \textbf{Acc. (\%)} \\
			\midrule
			\multirow{3}{*}{\shortstack{VGG13\\(74.64)}} & \multirow{3}{*}{\shortstack{VGG8\\(70.36)}} & AKD & 67.81 \\
			& & CTKD & 72.02 \\
			& & \textbf{KD + DTS} & \textbf{72.77} \\
			\midrule
			\multirow{3}{*}{\shortstack{ResNet50\\(79.34)}} & \multirow{3}{*}{\shortstack{MN-V2\\(64.60)}} & AKD & 53.86 \\
			& & CTKD & 63.17 \\
			& & \textbf{KD + DTS} & \textbf{63.24} \\
			\midrule
			\multirow{3}{*}{\shortstack{ResNet56\\(72.34)}} & \multirow{3}{*}{\shortstack{ResNet20\\(69.06)}} & AKD & 68.13 \\
			& & CTKD & 69.51 \\
			& & \textbf{KD + DTS} & \textbf{70.98} \\
			\midrule
			\multirow{3}{*}{\shortstack{ResNet110\\(74.31)}} & \multirow{3}{*}{\shortstack{ResNet32\\(71.14)}} & AKD & 70.34 \\
			& & CTKD & 72.05 \\
			& & \textbf{KD + DTS} & \textbf{72.20} \\
			\midrule
			\multirow{3}{*}{\shortstack{VGG13\\(74.64)}} & \multirow{3}{*}{\shortstack{MN-V2\\(64.60)}} & AKD & 52.53 \\
			& & CTKD & 62.28 \\
			& & \textbf{KD + DTS} & \textbf{62.44} \\
			\midrule
			\multirow{3}{*}{\shortstack{ResNet110\\(74.31)}} & \multirow{3}{*}{\shortstack{ResNet20\\(69.06)}} & AKD & 68.23 \\
			& & CTKD & 69.49 \\
			& & \textbf{KD + DTS} & \textbf{69.58} \\
			\bottomrule
		\end{tabular}
		
		\vspace{-0.5em}
	\end{table}
	
	\noindent\textbf{(b) Too high a temperature later on may lead to vanishing gradients.} In the late stage of training, student logits are often confident, with \( z_i \gg z_j \) for some \( i \). If we increase \( \mathcal{T} \to \infty \), the softmax becomes approximately uniform due to:
	\[
	\exp\left(\frac{z_i}{\mathcal{T}}\right) \approx 1 + \frac{z_i}{\mathcal{T}}, \quad \text{so} \quad P_{\text{S}, i}(\mathcal{T}) \to \frac{1}{C}.
	\]
	\indent Similarly, \( P_{\text{T}, i}(\mathcal{T}) \to \frac{1}{C} \). Substituting into the gradient:
	\[
	\frac{\partial \mathcal{L}_{\text{KD}}}{\partial z_i^{\text{S}}} = \frac{1}{\mathcal{T}} \left( P_{\text{S}, i}(\mathcal{T}) - P_{\text{T}, i}(\mathcal{T}) \right) \approx 0,
	\]
	\indent Because both distributions approach the uniform distribution as \( \frac{1}{\mathcal{T}} \to 0 \). The result is vanishing gradients, which: \textbf{(a)} Hinder precise refinement of the student model, \textbf{(b)} lead to underfitting despite prolonged training, and thus suppress beneficial guidance from the teacher. Therefore, using a high temperature in the late stages of training can nullify the learning signal, hampering final performance.
	
	The summary of this section can be stated as follows: (a) When \( \mathcal{T} \to 0 \) gradient magnitudes grow unboundedly, destabilizing training. (b) When \( \mathcal{T} \to \infty \); the gradients vanish, impeding convergence. This highlights the need to dynamically adjust temperature throughout training to maintain effective gradient flow.
	
	\begin{algorithm}[!t]
		\caption{Dynamic Temperature Scheduler (DTS)}
		\label{alg:dts_stabilized}
		\begin{algorithmic}[1]
			\Require 
			\Statex $\mathcal{T}_{\text{init}}$: Initial temperature
			\Statex $\mathcal{T}_{\text{min}}$, $\mathcal{T}_{\text{max}}$: Temperature bounds
			\Statex $e_{\text{current}}$, $e_{\text{max}}$: Current and total epochs
			\Statex $\mathcal{L}_t$, $\mathcal{L}_s$: Teacher/student CE losses
			\Statex $\mu$: Momentum coefficient (default: 0.9)
			\Ensure $\mathcal{T}_{\text{current}}$: Updated temperature
			
			\State Compute progress $p \gets e_{\text{current}} / e_{\text{max}}$
			\State $\mathcal{T}_{\text{cos}} \gets \mathcal{T}_{\text{init}} \cdot (0.5 + 0.5\cos(\pi p))$
			
			\State $d_{\text{loss}} \gets \mathcal{L}_s - \mathcal{L}_t$
			\State $\alpha \gets {d_{\text{loss}}}\text{/}({d_{\text{loss}}} + 1 + \epsilon)$ \Comment{$\epsilon$ prevents division by zero}
			
			\If{$\alpha > 1$}
			\State $\mathcal{T}_{\text{target}} \gets \mathcal{T}_{\text{init}} \cdot \mathcal{T}_{\text{cos}} \cdot \alpha$
			\Else
			\State $\mathcal{T}_{\text{target}} \gets \mathcal{T}_{\text{init}} \cdot \mathcal{T}_{\text{cos}}$
			\EndIf			\State $\mathcal{T}_{\text{target}} \gets \text{clip}(\mathcal{T}_{\text{target}}, \mathcal{T}_{\text{min}}, \mathcal{T}_{\text{max}})$
			
			\State $\mathcal{T}_{\text{current}} \gets \mu \cdot \mathcal{T}_{\text{current}} + (1-\mu)\cdot \mathcal{T}_{\text{target}}$
			\State \Return $\mathcal{T}_{\text{current}}$
		\end{algorithmic}
	\end{algorithm}
	
	\noindent\textbf{Architectural Gap.} When the architectural gap between the teacher and student models is large, the teacher tends to produce more extreme logits. In such cases, using a low temperature can exaggerate the mismatch between logits, making it harder for the student to align. Conversely, in settings where the architectures are similar, a high temperature may produce overly smoothed distributions, preventing the student from focusing on dominant classes. These insights align with the findings of Sun et al. \cite{logitstand} on the importance of logit standardization.
	
	\noindent\textbf{Empirical Evidence.} Studies such as CTKD, AKD, and RLKD have shown that adjusting the temperature leads to significant performance improvements over static temperature approaches. However, none of the previous works adjusted the temperature according to the student's need. Sun et al. \cite{logitstand} have also pointed out the inadequacy of having a fixed static temperature.%	shows that the larger models have logit output closer to zero mean and smaller standard variance, while the smaller models demonstrate larger standard variance on their logits
	
	\subsection{Dynamic Temperature Scheduler (DTS)}
	\label{DTS}
	\noindent Our temperature scheduler comprises three key components: (i) cosine scheduling, (ii) loss divergence-based adaptive scaling, and (iii) a smooth temperature update rule. These components are described in the following.
	
	\noindent\textbf{Cosine Scheduling.} The scheduler uses a modified cosine function to map training progress \( p \) to a value in the range \([0, 1]\). The training progress is defined as follows:
	\begin{equation}
		\label{eq:progress}
		p = \frac{e_c}{e_t},
	\end{equation}
	\indent where \( e_c \) is the current epoch, \( e_t \) is the total number of epochs, and \( p \in [0, 1] \) represents the fraction of training completed.
	\noindent The cosine scheduling curve \( S(p) \) is given by:
	
	\begin{equation}
		\label{eq:cosine_curve}
		S(p) = \lambda \cdot \left(1 + \cos(\pi \cdot p)\right).
	\end{equation}
	
	\indent We have set \(\lambda=0.5\) so, the formulation yields: \textbf{(a)} At the start of training: \( p = 0 \), \(S(0) = 0.5 \cdot \left(1 + \cos(0)\right) = 1.\) and \textbf{(b)} At the end of training: \( p = 1 \), \(S(1) = 0.5 \cdot \left(1 + \cos(\pi)\right) = 0.\). Hence, \( S(p) \) smoothly decreases from 1 at the beginning to 0 at the end of training.
	
	\begin{table}[!t]
		\centering
		\caption{Top-1 accuracy results of different KD methods paired with \textbf{our DTS} on same architecture teacher-student pairs on CIFAR-100. The best overall combination of logit distillation method is highlighted in \textcolor{red}{\textbf{RED}} and \textcolor{blue}{\textbf{Blue}} highlights the best gains \(\Delta\).}
		\label{tab:same_arch}
		\scriptsize
		\setlength{\tabcolsep}{2.0pt}
		\begin{tabular}{llcccccc}
			\toprule
			\textbf{Type} & \textbf{Method} & \multicolumn{6}{c}{\textbf{Model and Accuracy(\%)}} \\
			\midrule
			{Teacher} & & \shortstack{RNet32x4 \\{\scriptsize 79.42}} & \shortstack{RNet56 \\ \scriptsize72.34} & \shortstack{RNet110\\ \scriptsize 74.31} & \shortstack{WRN-40-2 \\ \scriptsize 75.61} & \shortstack{VGG13 \\ \scriptsize 74.64} & \shortstack{RNet110 \\ \scriptsize 74.31} \\
			\midrule
			{Student} & & \shortstack{ResN8x4\\{\scriptsize 72.50}} & \shortstack{ResN20\\{\scriptsize 69.06}} & \shortstack{ResN20\\{\scriptsize 69.06}} & \shortstack{WRN-16-2\\{\scriptsize 73.26}} & \shortstack{VGG8\\{\scriptsize 70.36}} & \shortstack{ResN32\\{\scriptsize 71.14}} \\
			\midrule
			\multirow{8}{*}{\rotatebox{90}{Feature}} 
			& FitNet & 73.50 & 69.21 & 68.99 & 73.58 & 71.02 & 71.06 \\
			& AT & 73.44 & 70.55 & 70.65 & 74.08 & 71.43 & 72.31 \\
			& RKD & 71.90 & 69.61 & 69.25 & 73.35 & 71.48 & 71.82 \\
			& CRD & 75.51 & 71.16 & 71.46 & 75.48 & 73.94 & 73.48 \\
			& OFD & 74.95 & 70.98 & 71.29 & 75.24 & 73.95 & 73.23 \\
			& ReviewKD & 75.63 & 71.89 & 71.34 & 76.12 & 74.84 & 73.89 \\
			& SimKD & 78.08 & 71.05 & 71.06 & 75.53 & 74.89 & 73.92 \\
			& CAT-KD & 76.91 & 71.62 & 71.37 & 75.60 & 74.65 & 73.62 \\
			\midrule
			\multirow{9}{*}{\rotatebox{90}{Logit}}
			& KD & 72.59 & 69.56 & 69.28 & 73.12 & 70.33 & 71.75 \\
			& \cellcolor{blue!10}KD + DTS & \cellcolor{blue!10}72.70 & \cellcolor{blue!10}70.31 & \cellcolor{blue!10}69.58 & \cellcolor{blue!10}73.63 & \cellcolor{blue!10}72.71 & \cellcolor{blue!10}\textcolor{red}{\textbf{72.20}} \\
			& \(\Delta\) & +0.11 & \textcolor{blue}{\textbf{+0.75}} & \textcolor{blue}{\textbf{+0.30}} & \textcolor{blue}{\textbf{+0.51}} & \textcolor{blue}{\textbf{+2.38}} & \textcolor{blue}{\textbf{+0.45}} \\
			\cmidrule{2-8}
			& DKD & 74.72 & 69.55 & 69.31 & 74.53 & 73.97 & 69.89 \\
			& \cellcolor{blue!10}DKD + DTS & \cellcolor{blue!10}74.73 & \cellcolor{blue!10}69.98 & \cellcolor{blue!10}69.43 & \cellcolor{blue!10}74.60 & \cellcolor{blue!10}\textcolor{red}{\textbf{74.03}} & \cellcolor{blue!10}70.27\\
			& \(\Delta\) & +0.01 & +0.43 & +0.12 & +0.07 & +0.06 & +0.38 \\
			\cmidrule{2-8}
			& MLKD & 75.03 & 70.85 & 70.32 & 75.18 & 70.52 & 70.32 \\
			& \cellcolor{blue!10}MLKD + DTS & \cellcolor{blue!10}\textcolor{red}{\textbf{75.25}} & \cellcolor{blue!10} \textcolor{red}{\textbf{70.98}} & \cellcolor{blue!10}\textcolor{red}{\textbf{70.58}} & \cellcolor{blue!10}\textcolor{red}{\textbf{75.58}} & \cellcolor{blue!10}{71.07} & \cellcolor{blue!10}70.54 \\
			& \(\Delta\) & \textcolor{blue}{\textbf{+0.22}} & +0.13 & +0.26 & +0.40 & +0.55 & +0.22 \\
			\bottomrule
		\end{tabular}
		\vspace{-0.5em}
	\end{table}
	
	\noindent\textbf{Loss Divergence and Adaptive Scaling.} To adjust the temperature based on the divergence between the teacher and student models, we first compute their cross-entropy losses:
	\begin{align}
		\mathcal{L}_t &= \text{CE}(\mathbf{z}_t, \mathbf{y}), \label{eq:teacher_loss} \\
		\mathcal{L}_s &= \text{CE}(\mathbf{z}_s, \mathbf{y}), \label{eq:student_loss}
	\end{align}
	\indent where \( \mathbf{z}_t \) and \( \mathbf{z}_s \) are the logits from the teacher and student models, respectively, \( \mathbf{y} \) is the ground-truth label, and \(\text{CE}(\cdot, \cdot)\) is the cross-entropy loss function.
	\noindent The loss divergence is then defined as:
	\begin{equation}
		\label{eq:loss_divergence}
		d_{\text{loss}} = \mathcal{L}_t - \mathcal{L}_s.
	\end{equation}
	\indent To ensure stability, we adaptively scale the divergence:
	\begin{equation}
		\label{eq:adaptive_scaling}
		\alpha = \frac{d_{\text{loss}}}{d_{\text{loss}} + 1 + \beta}.
	\end{equation}
	\indent Here, \( \beta \) is a very small integer to avoid division by 0. Note that at the beginning of training, when the student is untrained (\(\mathcal{L}_s > \mathcal{L}_t\)), \( d_{\text{loss}} < 0 \) so, \ref{eq:adaptive_scaling} yields positive result \(\alpha \). 
	\[
	\alpha = \frac{-d_{\text{loss}}}{-d_{\text{loss}} + 1 + \beta} > 0.
	\]
	
	\begin{table}[!t]
		\centering
		\renewcommand{\arraystretch}{0.9} % compact spacing
		\caption{Top-1 accuracy (\%) of different KD methods paired with \textbf{our DTS} across various teacher-student architectures on CIFAR-100. \textcolor{red}{\textbf{Red}} shows the best results; \textcolor{blue}{\textbf{Blue}} indicates improvements (\(\Delta\)) over baselines.}
		\label{tab:diff_arch}
		\begin{tabular}{cccccc}
			\toprule
			\textbf{Teacher} & \textbf{Student} & \textbf{Method} & \textbf{Acc. (\%)} & \textbf{\(\Delta\)}  \\
			\midrule
			\multirow{3}{*}{\makecell{ResNet32x4\\79.42}} & \multirow{3}{*}{\makecell{SHN-V2\\71.82}} & KD & 71.48 & –  \\
			& & KD + DTS & \textcolor{red}{\textbf{71.76}} & \textcolor{blue}{+0.28} \\
			& & DKD + DTS & 73.33 & \textcolor{blue}{–0.30} \\
			\midrule
			\multirow{3}{*}{\makecell{ResNet50\\79.34}} & \multirow{3}{*}{\makecell{MN-V2\\64.60}} & KD & 62.68 & –  \\
			& & KD + DTS & \textcolor{red}{\textbf{63.24}} & \textcolor{blue}{+0.56}  \\
			& & DKD + DTS & \textcolor{red}{\textbf{66.18}} & \textcolor{blue}{+0.03}  \\
			\midrule
			\multirow{3}{*}{\makecell{ResNet32x4\\79.42}} & \multirow{3}{*}{\makecell{VGG8\\70.36}} & KD & 69.95 & – \\
			& & KD + DTS & \textcolor{red}{\textbf{71.45}} & \textcolor{blue}{+1.50}  \\
			& & MLKD + DTS & \textcolor{red}{\textbf{73.35}} & \textcolor{blue}{+0.62}  \\
			\midrule
			\multirow{3}{*}{\makecell{WRN-40-2\\75.61}} & \multirow{3}{*}{\makecell{SHN-V2\\71.82}} & KD & 69.93 & –  \\
			& & KD + DTS & \textcolor{red}{\textbf{71.35}} & \textcolor{blue}{+1.42}  \\
			& & MLKD + DTS & \textcolor{red}{\textbf{74.24}} & \textcolor{blue}{+0.10}  \\
			\midrule
			\multirow{3}{*}{\makecell{VGG13\\74.64}} & \multirow{3}{*}{\makecell{MN-V2\\64.60}} & KD & 60.83 & – \\
			& & KD + DTS & \textcolor{red}{\textbf{62.44}} & \textcolor{blue}{+1.61} \\
			& & MLKD + DTS & \textcolor{red}{\textbf{66.74}} & \textcolor{blue}{+0.09} \\
			\bottomrule
		\end{tabular}
		% \vspace{-0.5em}
	\end{table}
	
	\begin{table}[!t]
		\footnotesize
		\tabcolsep=1.5mm
		\renewcommand\arraystretch{0.95} 
		\centering
		\caption{Top-1 accuracy (\%) between KD and KD with \textbf{DTS} on TinyImageNet. Best results are \textbf{bold}. Subtable (a) shows classical CNN-based models and (b) includes ViT-based models with ResNet-50 as the teacher.}
		\label{tab:tinyimagenet_combined}
		\begin{tabular}{clcccc}
			\toprule
			& \textbf{Teacher} & \textbf{Student} & \textbf{KD} & \textbf{KD + DTS} & \textbf{\(\mathcal{T}_{\text{max}} \rightarrow \mathcal{T}_{\text{min}}\)} \\
			\midrule
			\label{tab: tiny_image}
			\multirow{2}{*}{(a)} & ResNet34 & ResNet18 & 57.82 & \textbf{60.75} & 4$\rightarrow$2 \\
			& ResNet50 & MN-V1 & 56.84 & \textbf{61.18} & 4$\rightarrow$2 \\
			\midrule
			\label{tab: vit}
			\multirow{4}{*}{(b)} & \multirow{4}{*}{ResNet50} & DeiT-Ti & 68.42 & \textbf{68.58} & 4$\rightarrow$2 \\
			& & T2T-ViT-7 & 64.03 & \textbf{64.60} & 4$\rightarrow$2 \\
			& & PvT-Ti & 72.01 & \textbf{72.69} & 3$\rightarrow$1 \\
			& & PiT-Ti & 74.04 & \textbf{74.18} & 3$\rightarrow$1 \\
			\bottomrule
		\end{tabular}
		\vspace{-0.5em}
	\end{table}
	
	\subsection{Temperature Update}
	\noindent The temperature \(\mathcal{T}\) is updated dynamically in two steps: (a) Clamped temperature and (b) Smooth update.
	
	\noindent \textbf{(a) Clamped Temperature:} Compute the tentative temperature using:
	\begin{equation}
		\label{eq:clamped_temp}
		\mathcal{T}_{\text{clamped}} = \text{clamp}(x, a, b)
	\end{equation}
	
	where, \(x =\mathcal{T}_{\text{init}} \cdot S(p) \cdot \alpha\), \(a=\mathcal{T}_{\text{min}} \) and \(b = \mathcal{T}_{\text{max}} \) \(\mathcal{T}_{\text{init}}\) is the initial (base) temperature. \(S(p)\) Eq. \ref{eq:cosine_curve} is a scaling function of progress \(p \in [0, 1]\), \(\alpha\) Eq. \ref{eq:adaptive_scaling} is an modulation coefficient controlling the aggressiveness of scaling, \(\mathcal{T}_{\text{min}}\) and \(\mathcal{T}_{\text{max}}\) define the allowable bounds for the temperature, \(\text{clamp}(x, a, b)\) restricts the value of \(x\) to the closed interval \([a, b]\). We have further discussed about this clamping in Section~\ref{ablation}.
	
	\noindent \textbf{(b) Smooth Update:} Update the current temperature using a momentum-based approach:
	\begin{equation}
		\label{eq:smooth_update}
		\mathcal{T}_{\text{current}} = \mu \cdot \mathcal{T}_{\text{prev}} + (1 - \mu) \cdot \mathcal{T}_{\text{clamped}}
	\end{equation}
	
	where \(\mu \in [0, 1]\) is the momentum term and \(\mathcal{T}_{\text{prev}}\) is the temperature from the previous epoch. This smooths the temperature transitions to avoid abrupt changes during training.

    \begin{table*}[!th]
		\caption{Evaluation results of \textbf{our DTS}. Only Rouge-L (R-L) scores are shown. The best values in each column block are \textbf{bolded}. The supervised fine tuning (SFT w/o KD) was done on 20 epoch but the rest was trained on 7 epochs.}
		\centering
		\small
		\setlength{\tabcolsep}{10pt}
		\begin{tabular}{lrl|ccccc}
			\toprule
			\multirow{2}{*}{Model} & \multirow{2}{*}{Params} & \multirow{2}{*}{Method}
			& Dolly & Self & S-NI & UnNI & Vicuna \\
			& & & R-L & R-L & R-L & R-L & R-L \\ 
			\midrule
			\multirow{6}{*}{\rotatebox{90}{GPT-2}} 
			& 1.5B & \cellcolor{gray!10}Teacher & \cellcolor{gray!10}27.6 & \cellcolor{gray!10}14.3 & \cellcolor{gray!10}27.6 & \cellcolor{gray!10}31.8 & \cellcolor{gray!10}16.3 \\ 
			\cmidrule(l){2-8}
			& \multirow{5}{*}{120M} 
			& SFT w/o KD & 23.3 & 10.0 & 14.7 & 18.5 & 16.3 \\
			& & KD & 21.88 & 9.74 & 14.88 & 18.46 & 13.92 \\
			& & \cellcolor{blue!10}KD + DTS & \cellcolor{blue!10}21.48 & \cellcolor{blue!10}9.82 & \cellcolor{blue!10}\textbf{16.16} & \cellcolor{blue!10}\textbf{19.66} & \cellcolor{blue!10}13.95 \\
			& & SeqKD & \textbf{22.27} & 9.47 & 14.83 & 17.37 & 14.90 \\
			& & \cellcolor{blue!10}SeqKD + DTS & \cellcolor{blue!10}21.30 & \cellcolor{blue!10}\textbf{10.89} & \cellcolor{blue!10}15.54 & \cellcolor{blue!10}18.77 & \cellcolor{blue!10}14.15 \\ 
			\midrule
			\multirow{6}{*}{\rotatebox{90}{OPT}} 
			& 1.3B & \cellcolor{gray!10}Teacher & \cellcolor{gray!10}26.0 & \cellcolor{gray!10}11.4 & \cellcolor{gray!10}23.1 & \cellcolor{gray!10}28.4 & \cellcolor{gray!10}15.6 \\ 
			\cmidrule(l){2-8}
			& \multirow{5}{*}{125M}
			& SFT w/o KD & -- & -- & -- & -- & -- \\
			& & KD & 19.41 & 8.06 & 15.30 & 17.52 & 13.65 \\
			& & \cellcolor{blue!10}KD + DTS & \cellcolor{blue!10}19.91 & \cellcolor{blue!10}8.46 & \cellcolor{blue!10}15.64 & \cellcolor{blue!10}\textbf{19.66} & \cellcolor{blue!10}14.05 \\
			& & SeqKD & 20.14 & 8.24 & 15.52 & 17.98 & \textbf{14.64} \\
			& & \cellcolor{blue!10}SeqKD + DTS & \cellcolor{blue!10}\textbf{20.33} & \cellcolor{blue!10}\textbf{8.95} & \cellcolor{blue!10}\textbf{16.74} & \cellcolor{blue!10}18.25 & \cellcolor{blue!10}13.58 \\
			\bottomrule
		\end{tabular}
		% \vspace{-0.5em}
		\label{tab: llmEval}
	\end{table*}
	
	\begin{table}[!t]
		\caption{Performance of \textbf{TinyBERT-4L-312D (Student)} distilled from \textbf{BERT\_base\_uncased (Teacher)} on GLUE tasks. \textbf{OE} = Our Embedding layer, \textbf{OP} = Our Prediction layer.}
		\label{tab: bertGlueDistill}
		\centering
		\footnotesize
		\resizebox{0.5\textwidth}{!}{%
			\begin{tabular}{lccc}
				\toprule
				\textbf{Method} & \textbf{CoLA (MCC)} & \textbf{MRPC (Acc)} & \textbf{RTE (Acc)} \\
				\midrule
				SFT & 0.52 & 87.00 & 72.00 \\
				\midrule
				TinyBERT & 0.35 & 83.57 & 67.14 \\
				\rowcolor{blue!10}  with OP & 0.36 & 84.31 & \textbf{67.87} \\
				\rowcolor{blue!10}  OE + OP & \textbf{0.39} & \textbf{84.33} & 66.06 \\
				\bottomrule
		\end{tabular}}
		\vspace{-0.5em}
	\end{table}
	
	\section{Experiment}
	\subsection{Setup}
	\noindent\textbf{Models.} For computer vision tasks, we used both CNN-based (e.g., ResNet \cite{resnet}, VGG \cite{vgg}) and Transformer-based (ViT \cite{vit}) models (e.g., PiT \cite{pit} , DeiT \cite{deit}). For natural language processing tasks, we use Transformer-based models including BERT \cite{bert}, TinyBERT \cite{tinybert}, and generative models GPT-2 \cite{gpt2} and OPT \cite{opt}.
	
	\noindent\textbf{Datasets.} For computer vision tasks, we use CIFAR-100 \cite{cifar}, a labeled dataset with 100 classes, and Tiny-ImageNet \cite{tinyimagenet}, a subset of ImageNet-1K \cite{russakovsky2015imagenet} containing 200 classes. For the BERT model experiments, we use 3 of the 9 GLUE \cite{glue} benchmark tasks: \textbf{CoLA}, \textbf{MRPC}, and \textbf{RTE}. The generative language models are trained on the Dolly-15k dataset \cite{dolly} and evaluated on \textbf{DollyEval}, \textbf{SelfInst} \cite{self_inst}, \textbf{VicunaEval} \cite{vicuna}, \textbf{S-NI} \cite{super-natural-instructions}, and \textbf{UnNI} \cite{uinst}. We use the ROUGE-L (R-L) score \cite{lin-2004-rouge} to assess generation quality.
	
	\noindent\textbf{Implementation Details.} For computer vision, all models are trained using SGD \cite{sgd} optimizer. For CIFAR-100, training runs for 100 epochs, the initial learning rate is 0.01 for MobileNets and ShuffleNets, and 0.1 for other models with decay by 0.1 at the 63rd, 87th, and 92nd epochs. For Tiny-ImageNet, training runs for 50 epochs, initial learning rate was 0.2, decayed by 10$10\times$ at epochs 15, 30, and 45. Results are averaged over 3 trials. All hyperparameters are adopted from corresponding baselines, with the exception of training epochs. Temperature for nlp models: \(\mathcal{T}_\text{init} = \mathcal{T}_\text{max} = 4\), \(\mathcal{T}_\text{min} = 2\).
	
	\subsection{Results and Analysis.}
	\noindent In this section, we provide a detailed overview of the evaluation of our DTS through computer vision and natural language processing tasks. 
	For computer vision tasks, in Tab. \ref{tab:diff_scheduler}, we demonstrate the effectiveness of our scheduler by comparing it with AKD \cite{AKD} and CTKD \cite{CTKD} paired with vanilla KD \cite{kd}. Our method outperformed other scheduling techniques in all our tests. Then we also tested our model, pairing it with other KD methods as shown in Tab. \ref{tab:same_arch} and \ref{tab:diff_arch}, and achieved improved results in almost all tests, especially when paired with vanilla KD.
	We then apply our scheduler to natural language processing tasks as shown in Tab.~\ref{tab: bertGlueDistill} and \ref{tab: llmEval}; in all cases DTS improves the existing NLP distillation methods. 
	Overall, these results validate our scheduler to be extremely versatile. From analyzing the results, our observations are: (1) Our scheduler heavily depends on the scheduling range and (2) Selection of the range is also dictated by the architectural gaps of the models.
	
	\subsection{Ablation}
	\label{ablation}
	\begin{figure}[t]
		\centering
		\begin{tikzpicture}
			\begin{axis}[
				width=0.95\linewidth,
				height=4cm,
				ylabel={Accuracy (\%)},
				symbolic x coords={3$\rightarrow$1,4$\rightarrow$2,6$\rightarrow$4,8$\rightarrow$4,11$\rightarrow$9},
				xtick=data,
				ymin=55, ymax=61,
				legend style={at={(0.8,0.3)}, anchor=north, legend columns=-1},
				ymajorgrids=true,
				xmajorgrids=true,
				grid style=dashed,
				]
				
				% ResNet56
				\addplot[
				color=blue,
				mark=square*,
				line width=1pt,
				]
				coordinates {
					(3$\rightarrow$1,58.94)
					(4$\rightarrow$2,57.92)
					(6$\rightarrow$4,58.36)
					(8$\rightarrow$4,59.34)
					(11$\rightarrow$9,60.21)
				};
				
				% ResNet110
				\addplot[
				color=red,
				mark=*,
				line width=1pt,
				]
				coordinates {
					(3$\rightarrow$1,57.49)
					(4$\rightarrow$2,57.05)
					(6$\rightarrow$4,57.87)
					(8$\rightarrow$4,57.47)
					(11$\rightarrow$9,58.52)
				};
				
				\legend{56, 110}
			\end{axis}
		\end{tikzpicture}
		\caption{This figure shows the performance of the ResNet20 distilled by \textcolor{blue}{ResNet56} and \textcolor{red}{ResNet110} on a 50 epoch cifar-100 training on different ranges of temperature, without modifying optimizer settings. The horizontal x-axis represents the ranges of temperatures.}
		\label{fig: ablation}
		
		\vspace{-0.5em}
	\end{figure}
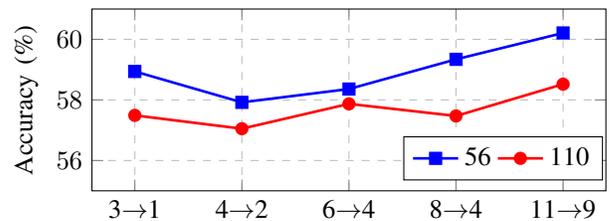
	\noindent For ablation, we conducted extensive experiments with different temperature ranges to determine which range performs better. In our experiments, we found that the range from \( \mathcal{T}_{\text{max}} = 8\) to \( \mathcal{T}_{\text{min}} = 4\) yields good results for same-architecture distillation (Tab.~\ref{tab:same_arch}); thus, we use this range for all of our experiments on CIFAR-100 benchmarking. However, we found that a lower temperature range (\( \mathcal{T}_{\text{max}} = 3\) to \( \mathcal{T}_{\text{min}} = 1\)) helped with cross-architecture distillation (Tab.~\ref{tab:diff_arch}). To demonstrate how different ranges of temperature affect performance, we show an example in Fig.~\ref{fig: ablation}. For the adaptive scaling method (Eq.~\ref{eq:adaptive_scaling}), in our tests, if we do not use the clamping method described in Eq.~\ref{eq:clamped_temp}, the temperature changes abruptly, causing the student model to perform even worse.

    \textbf{Limitations.} Our method has several limitations. First, although we dynamically adjust the temperature, the student and teacher still share the same temperature value. Second, the selection of the temperature range (\(\mathcal{T}_{\text{max}}\) to \(\mathcal{T}_{\text{min}}\)) requires tuning, as demonstrated in Fig.~\ref{fig: ablation}, where different temperature ranges yield varying effects depending on the training epoch. Third, we adjust the temperature batch-wise, which may be suboptimal; particularly challenging batches can destabilize the training process and hinder convergence.
	
	\section{Conclusion}
	\noindent In this paper, we identify the drawback of having a static temperature throughout the training process and highlight the importance of a temperature scheduling mechanism for KD pipelines to regulate the temperature. We introduce a novel dynamic temperature scheduler, which dynamically assigns temperature during the training process according to the teacher-student output difference. In all of our conducted tests, on both computer vision tasks and natural language processing tasks, our scheduler improved and outperformed existing KD methods.
	
	\textbf{Future Work.} Several promising directions remain for future investigation. First, assigning different temperatures to the student and teacher models separately may help bridge the performance gap by caused by architectural gap. Second, instance-wise temperature adjustment could further improve distillation by adapting the temperature based on sample difficulty, allowing easier samples to use lower temperatures while challenging samples benefit from higher temperatures.
	
	\section*{Acknowledgements}
	\noindent We would like to express our sincere gratitude to Mustavi Rahman Khan and Jubayer Ibn Hamid for their continuous support, extremely valuable feedback, and encouragement throughout the preparation of this paper.
	
	\bibliographystyle{IEEEtran}
    \bibliography{ref}
\end{document}